\documentclass{article}

\usepackage[preprint,nonatbib]{neurips_2025}

\usepackage[utf8]{inputenc} 
\usepackage[T1]{fontenc}    
\usepackage{hyperref}       
\usepackage{url}            
\usepackage{booktabs}       
\usepackage{amsfonts}       
\usepackage{nicefrac}       
\usepackage{microtype}      
\usepackage{xcolor}         

\usepackage{amsmath}
\usepackage{amssymb}
\usepackage{mathtools}
\usepackage{amsthm}

\usepackage{array}  
\usepackage{bm}
\usepackage{graphicx}
\usepackage{caption} 
\usepackage{multirow}

\definecolor{darkblue}{RGB}{0,0,139}
\definecolor{darkred}{RGB}{139,0,0}

\theoremstyle{plain}
\newtheorem{theorem}{Theorem}[section]

\theoremstyle{definition}
\newtheorem{definition}[theorem]{Definition}

\theoremstyle{remark}

\title{Position: Foundation Models Need Digital Twin Representations}

\author{
Yiqing Shen, Hao Ding, Lalithkumar Seenivasan, Tianmin Shu, Mathias Unberath\\
Department of Computer Science Johns Hopkins University\\
\texttt{yshen92@jhu.edu}
}

\begin{document}

\maketitle

\begin{abstract}
Current foundation models (FMs) rely on token representations that directly fragment continuous real-world multimodal data into discrete tokens.
They limit FMs to learning real-world knowledge and relationships purely through statistical correlation rather than leveraging explicit domain knowledge. 
Consequently, current FMs struggle with maintaining semantic coherence across modalities, capturing fine-grained spatial-temporal dynamics, and performing causal reasoning. 
These limitations cannot be overcome by simply scaling up model size or expanding datasets. 
This position paper argues that the machine learning community should consider digital twin (DT) representations, which are outcome-driven digital representations that serve as building blocks for creating virtual replicas of physical processes, as an alternative to the token representation for building FMs. 
Finally, we discuss how DT representations can address these challenges by providing physically grounded representations that explicitly encode domain knowledge and preserve the continuous nature of real-world processes. 
\end{abstract}

\section{Introduction}
Current multimodal foundation models (FMs), such as multimodal large language models (LLMs), have already demonstrated their capabilities in processing and understanding diverse modalities like text, images, and videos \cite{wang2024mio,wang2024survey}. 
These FMs typically rely on token representations \cite{tan2024sweettokenizer}, which discretize continuous multimodal inputs into tokens representing image patches, or video frames. 
It therefore introduces various limitations that can constrain FMs' ability to achieve genuine multimodal understanding and reasoning  \cite{li2024multimodal,wang2024comprehensive}.

The primary issue lies in how token representations fragment the natural continuity of real-world knowledge and relationships \cite{zhan2024anygpt}. 
Specifically, when processing multimodal data, these representations create artificial boundaries between modalities, sacrificing important physical relationships and semantic connections \cite{wu2024towards}.
For instance, in video analysis, discretizing continuous visual and temporal information into fixed-length token sequences compromises the fine-grained spatial relationships and temporal dynamics essential for understanding complex scenarios \cite{wu2024smart}.
This discretization forces FMs to learn fundamental physical principles and relationships purely through statistical correlations, rather than leveraging explicit physical constraints and domain knowledge.
Moreover, token representations struggle to maintain semantic coherence across modalities \cite{wu2024semantic}.
Without an explicit framework for encoding real-world knowledge such as physical laws and domain constraints that govern real-world interactions, current FMs suffer from capturing complex causal relationships \cite{khan2023drawbacks}. 
This limitation particularly affects their ability to perform counterfactual reasoning and generate physically plausible predictions, making them susceptible to spurious correlations and limiting their generalization capabilities \cite{zhao2020training}.

While previous efforts to address these limitations have focused on scaling up model size and expanding training datasets \cite{chen2024expanding}, they face inherent constraints due to data availability and computational costs. 
Even synthetic data generation methods can struggle to produce training examples that adequately reflect real-world physics and domain-specific constraints \cite{hao2024synthetic}.
This misalignment occurs because token representations lack explicit modeling of physical laws and domain rules, resulting in generated data that inadequately captures the complexities of real-world scenarios.
These challenges indicate the need for an alternative representation paradigm for FMs that better captures the inherent structure and relationships in multimodal data \cite{aasi2024generating,deiseroth2024t}. 
This paradigm is expected to preserve rich semantic relationships across modalities while enabling more efficient and interpretable model operations that maintain alignment with physical constraints and domain knowledge \cite{manzoor2023multimodality}.

Digital twins (DTs) are outcome-driven replicas of physical processes, capturing and modeling task-specific entities and interactions to analyze and optimize the physical counterpart.
%
While traditional DTs mostly focus on industrial applications like manufacturing, smart cities, and healthcare \cite{vallee2023digital}, we propose extending this concept to create universal semantic representations for FMs. 
DT representations are the building blocks of DTs, comprising outcome-driven digital representations extracted from raw data. 
Unlike token representations that fragment continuous information through direct vectorization, DT representations preserve the semantic and physical relationships inherent in multimodal data while maintaining domain-specific constraints \cite{dihan2024digital,yang2021developments}.
The DT paradigm also introduces an architectural separation between low-level representation extraction and high-level FM analysis.
Real-world data undergoes outcome-driven processing to extract meaningful representations that capture task-specific entities and interactions, which maps diverse multimodal inputs into unified, interpretable formats while preserving essential semantic relationships.
This preservation therefore enables FMs to operate on representations that directly reflect real-world properties and relationships.


\paragraph{Our Position}
%
The machine learning community should more seriously consider alternative representations of multimodal data beyond tokens. 
%
Token representations have been demonstrated to exhibit inherent limitations that cannot be overcome by simply scaling up model size or expanding datasets. 
We posit that \textbf{alternative representations, such as DT representations, offer a promising paradigm shift. 
These representations will enable multimodal FMs to capture the continuous nature of real-world information better, preserve physical constraints, and facilitate meaningful cross-modal interactions}.
This shift from statistical pattern matching to physically grounded representations will enhance model capabilities across multiple dimensions \textit{i}.\textit{e}., from improved generalization and interpretability to more efficient scaling. 

\paragraph{Alternative Views}
Several plausible counterarguments contrast our position on adopting DT representations for FMs. 
First, proponents of token representations argue that the success of current FMs demonstrates the effectiveness of learned statistical representations \cite{viswanathan2025geometry}. 
They contend that token representations, despite their limitations, have proven capable of capturing complex patterns and relationships across modalities without requiring explicit physical modeling \cite{chen2024multi}. 
The historic improvements in performance through architectural innovations and scaling suggest that token representations may eventually overcome their current limitations \cite{bigverdi2024perception,wang2022multimodal}.
Another perspective criticizes that the computational overhead of maintaining and processing DT representations could outweigh their benefits. 
These challenges may be most pronounced in diverse real-world scenarios, where token representations retain their simplicity. 
%
%
%
They suggest that improving existing token representations through better architectures and training methods might be more cost-effective than developing entirely new representation paradigms \cite{li2024multi}.
Some researchers also question whether explicit physical modeling is necessary for achieving robust multimodal understanding \cite{vaidya2021neural}. 
They argue that human cognition often operates on abstract representations rather than detailed physical models, and FMs might similarly develop effective reasoning capabilities through learned abstractions rather than explicit physical constraints \cite{courellis2024abstract}. 
This view suggests that the limitations of current token representation might be addressed through improved training objectives and architectural designs rather than fundamental changes to the representation strategy \cite{shen2023movit,choi2025emerging}.
These alternative perspectives highlight important considerations about the practicality, efficiency, and necessity of transitioning to DT representations. 
%

\section{Current Landscape of Foundation Models}

\paragraph{Foundation Models}
%
%
%
Containing billions or trillions of parameters, FMs usually learn generalizable representations through a self-supervised manner on massive datasets, enabling them to transcend their initial training domains \cite{moor2023foundation, awais2025foundation}.
Furthermore, FMs generally offer the flexibility to be adapted to specific datasets and novel downstream tasks in a few-shot (\textit{e}.\textit{g}., fine-tuning) or a zero-shot (\textit{e}.\textit{g}., prompt engineering) manner \cite{liu2024few}. 
Such transformative potential of FMs emerged first in the domain of natural language processing, where Transformer-based large language models (LLMs) demonstrated unprecedented zero-shot capabilities (\textit{i}.\textit{e}., the ability to handle novel tasks without task-specific training) in various tasks ranging from text generation, summarization, and translation. 
This zero-shot generalization capability has now been expanded into multimodal domains, with recent FMs processing combinations of text, images, audio, and video through unified architectures.

%


\paragraph{Token Representations}
Transformers are the building blocks of most FMs \cite{sharir2021image}. 
Initially proposed for text processing, Transformers employ tokenizers to convert raw text into token representations for the model to process \cite{devlin2018bert}. 
These tokens then go through an embedding layer, transforming each token into a continuous vector. 
%
%
Furthermore, depending on the specific application, tokens may further undergo positional embedding, to embed the order of tokens or preserve spatial coordinates. 
With the expansion of token representation to other modalities, it emerged as a flexible and common representation among varied modalities, enabling FMs to process modalities other than text, such as image, video, and audio \cite{zhang2024word}. 
The common token representations among multiple modalities and the inclusion of token-type embedding layers, further enabled FMs to process multi-modal data, enabling multi-modality downstream applications.


\paragraph{Token Representation for Text}
Raw text data, being unstructured, cannot be directly processed by FMs at scale. 
Tokenizing text thus becomes a fundamental step in converting raw text data into a structured, numerical format that the models can process \cite{devlin2018bert,sharir2021image}. 
In FMs, depending on the desired wide range of downstream applications and the backbone model architecture, different types of text tokenizers are employed to break down a text into a series of words or sub-words, which are then mapped to IDs based on pre-defined vocabularies. 
Different types of tokenizers include (a) word-level tokenizers, (b) sub-word tokenizers such as (i) Byte-Pair Encoding (BPE), (ii) WordPiece, and (iii) SentencePiece, (c) Charecter-level tokenizers, (d) Byte-level tokenizers and (e) Sentence-level tokenizers. 
The most widely adopted tokenizers are (a) BPE used in GPT-2, (b) WorkPiece tokenizer employed in BERT, Byte-level BPE used in RoBERTa, hybrid Byte-level BPE tokenizer used in GPT-3 and GPT-4 \cite{devlin2018bert}. 
Upon converting a text into a series of tokens, they undergo embedding -- encompassing token embedding, position embedding, and token type embedding \textit{i}.\textit{e}., mapping discrete tokens into dense, continuous vectors, capturing semantic and syntactic information.

\paragraph{Token Representation for Image}
Establishing a uniform data representation for Transformers to seamlessly process images, image tokenization transforms raw image data into a sequence of image tokens \cite{sharir2021image}. 
Widely employed image tokenization techniques include (a) Patch-based tokenization, (b) object-based tokenization, (c) super-pixel or region-based tokenization, and (d) pixel-based tokenization. 
In patch-based tokenization, a raw image is divided into $n$ number of uniform patches \cite{sharir2021image,ronen2023vision}. 
In the object-based tokenization, patches with objects of interest are extracted from raw image data using object detection and region proposal networks \cite{chen2024subobject}. 
Super-pixel or region-based tokenization employs segmentation models to segment regions of interest based on colors, textures, or objects, and generate patches from raw images accordingly \cite{lew2024superpixel}. 
Pixel-based tokenization is a rare method in which each pixel in the image data is treated as a token. 
In the first three tokenization methods, tokens can be embedded with feature vectors usually extracted from the patches by convolutional neural networks (CNNs). 
%
%
In pixel-based tokenization, tokens (pixels) are embedded by directly encoding the pixel values into feature vectors.

\paragraph{Token Representation for Video}
With images (frames) being the building blocks of video data, most video tokenization methods leveraged and expanded image tokenization approaches \cite{wang2024omnitokenizer}. 
These include (a) frame-based tokenization, (b) key-frame-based tokenization, and (c) object-based tokenization. 
In frame-based tokenization, each frame is considered as a token. 
Considering the possible lack of distinct features between consecutive frames and the limited scalability of compute resources for large videos, key-frame-based tokenization is employed, where key-frames selected based on pre-defined conditions are considered as tokens \cite{liang2024keyvideollm}. 
Object-based tokenization can be employed, where regions with objects of interest in every/key-frames are extracted and used as tokens \cite{tian2024tokenize}. 
%
%
Alternate to these methods, spatial-temporal patch tokenization uses 3D patches, with each patch holding both spatial and temporal information \cite{jang2024efficient,ryoo2021tokenlearner}.





\section{Defining and Formalizing Digital Twin Representations}

\begin{figure*}[t]
    \centering
    \includegraphics[width=\linewidth]{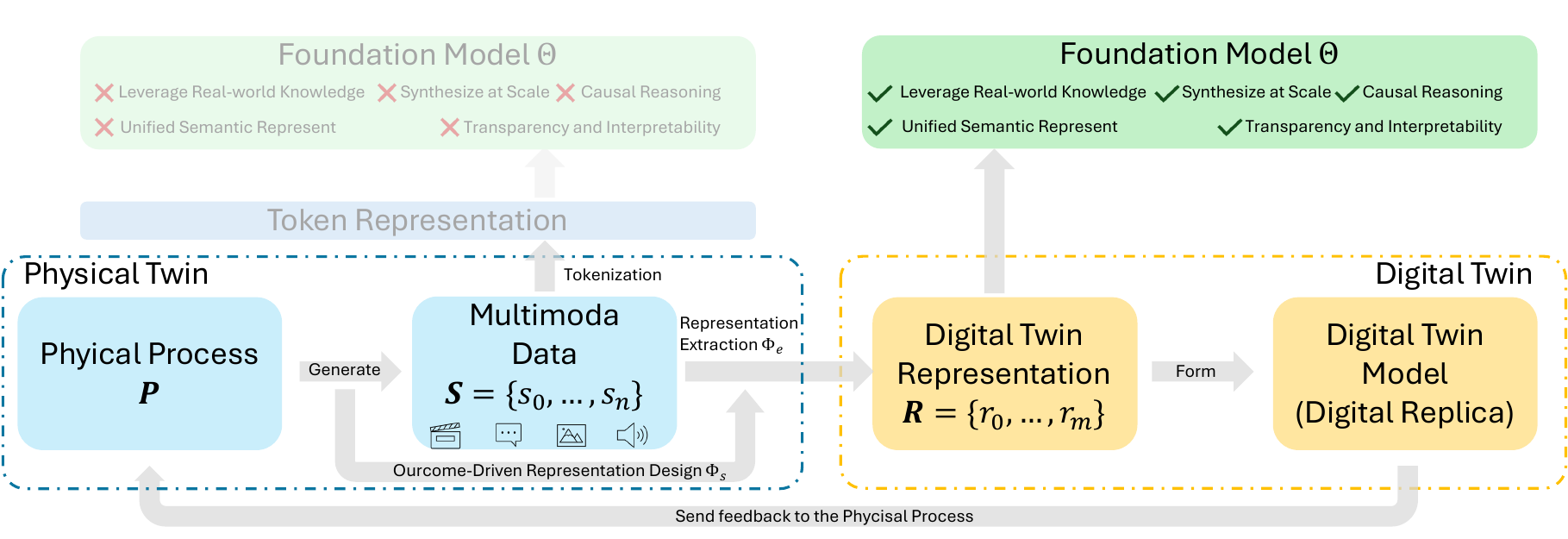}
    \caption{The illustration of the DT paradigm as well as the difference and relation between DT representations and token representations.}
    \label{fig:digital_twin}
\end{figure*}

\subsection{Definition}
To establish our position on DT representations for FMs, we must first clearly define digital twins (DTs).
DTs have transformed numerous domains, from smart cities~\cite{mohammadi2017smart, brandtstaedter2018digital, chen2018digital} to manufacturing~\cite{bilberg2019digital, mandolla2019building}, and healthcare~\cite{shu2023twin, ding2024digital, ding2024towardsS, ding2024towardsR}. 
Fuller \textit{et al.} \cite{fuller2020digital} reviewed several definitions \cite{glaessgen2012digital, chen2017integrated, liu2018role, zheng2019application, erkoyuncu2018digital, madni2019leveraging} of the DT.
For example, NASA~\cite{glaessgen2012digital} stated that ``\textit{A digital twin is an integrated multiphysics, multiscale, probabilistic simulation of an as-built vehicle or system that uses the best available physical models, sensor updates, fleet history, etc., to mirror the life of its corresponding flying twin.}"; 
Zheng \textit{et al.}~\cite{zheng2019application} stated ``\textit{A digital twin is a set of virtual information that fully describes a potential or actual physical production from the micro atomic level to the macro geometrical level.}";
Madni stated ``\textit{a digital twin is a virtual instance of a physical system (twin) that is continually updated with the latter’s performance, maintenance, and health status data throughout the
physical system’s life cycle.}".
Although the definitions vary, they all define DT in an outcome-driven manner. 
This means the DT representation should also be crafted according to the demand of modeling the task-specific modalities and interactions. 
Following Fuller \textit{et al.}~\cite{fuller2020digital}, we emphasize that a DT should facilitate automatic cyclic data flows between the physical object and its digital counterpart. 
This distinguishes them from other concepts like digital models and digital shadows where there is either no data flow or only one-way data flow from physical object to digital. 
Based on these insights, we propose the following definition for DT:
\begin{definition}
\label{def:digital_twin}
A digital twin is an outcome-driven paradigm that creates and maintains a dynamic digital replica of a physical process, capturing and modeling task-specific entities and interactions to analyze and optimize the physical counterpart.
\end{definition}

Such DT paradigm is depicted in Fig.~\ref{def:digital_twin}. 
We described the data flow from the physical process to the DT in more detail to clarify the definition of DT representation. 
The data is first captured by physical sensors in the form of raw data, \textit{e}.\textit{g}. images and robot kinematics, and then goes through an outcome-driven processing to form DT representations. 
The DT representations are the building blocks of the DT through mixed reality technologies like simulation or augmented reality. 
According to the DT paradigm, we define DT representations as follows:
\begin{definition}
\label{def:digital_twin_representation}
The digital twin representations are the set of outcome-driven digital representations extracted from raw data serving as building blocks of the digital twin model in the digital twin paradigm.
\end{definition}


\paragraph{Relationship with Neural Network Latent Spaces}
DT representations and neural network latent spaces can be complementary rather than mutually exclusive. 
When designed to capture task-specific entities and interactions, latent space representations can serve as components of DT representations. 
The major distinction lies in the outcome-driven nature of DT representations \textit{i}.\textit{e}., they are explicitly constructed to model specific aspects of the physical process rather than emerging solely from general-purpose training objectives.
%


\paragraph{Distinction from Token Representations}
Token representations fundamentally differ from DT representations in their construction and purpose. 
To be more specific, token representations are generated through a direct vectorization of raw data and embedded into a latent space, which is therefore not outcome-driven.
Consequently, by design, token representations do not necessarily capture task-specific entities or processes. 
For example, the patch embedding in ViT as the perception module is generated on a uniformly divided patch, where one patch may contain no entities of interest or multiple entities of interest in the task. 
In an outcome-driven manner, the DT representations should consist of features of task-specific entities like cars and traffic lights as well as the interaction.

\subsection{Attempted Formalization}
To advance our position, we propose an attempted formalization of DT representation-based FM development.
Let $\mathbf{P}$ represent a physical process for which we aim to develop an FM ($\mathbf{\Theta}$) that produces target outputs ($\mathbf{T}$). 
The FM operates on a set of DT representations $\mathbf{R} = \{r_0, r_1, ..., r_m\}$ where $r_i$ is one modality of outcome-driven digital representation of the physical process that encodes task-specific properties, such as identification, geometric property, or physical constraint of entities of interest or interactions among them, expressed as $\mathbf{T} = \mathbf{\Theta}(\mathbf{R})$.
These DT representations are derived from raw sensor data $\mathbf{S} = \{s_0, s_1, ..., s_n\}$ through an outcome-driven processing pipeline (\textit{i}.\textit{e}., DT construction process) $\mathbf{\Phi}$. 
Here, $s_i$ denotes one modality of a measurement or observation from physical sensors, such as images, depth maps, or kinematic data.
This processing $\mathbf{R} = \mathbf{\Phi}(\mathbf{P}, \mathbf{S})$ consists of two main steps \textit{i}.\textit{e}., (1) outcome-driven representation design $\mathbf{\Phi}_d$ and (2) representation extraction $\mathbf{\Phi}_e$. 
Outcome-driven representation design $\mathbf{\Phi}_d$ is a design process that decides the forms and components of the representation according to the task-specific requirement. 
The representation extraction $\mathbf{\Phi}_e$ is to use processing algorithms to extract specific representation from the raw sensor data. 
The process can be represented as $\mathbf{R} = \mathbf{\Phi}(\mathbf{P}, \mathbf{S}) = \mathbf{\Phi}_e(S, \mathbf{\Phi}_d(\mathbf{P}))$. 

\section{Advantages of Digital Twin Representations}

FMs building upon current token representation can face various limitations in processing and understanding multimodal data, which manifest across multiple dimensions:
(1) the ability to leverage existing domain knowledge, 
(2) the capacity to generate high-quality synthetic data at scale, 
(3) the capability to perform robust causal reasoning, 
(4) the effectiveness of semantic representations across modalities, 
and (5) the degree of transparency and interpretability. 
In this section, we present five assumptions about how DT representations can address these challenges, offering a promising alternative paradigm that could transform how FMs interact with and reason about the real world. 
%

\subsection{Assumption 1: DT Representation Characterizes Real-World Knowledge and Relationships Better}
Understanding real-world knowledge and relationships such as semantic properties, geometric relationships, physical laws, and \textit{etc} enables FMs to achieve meaningful interaction and reasoning \cite{gupta2024essential}. 
Yet, current FMs with token representations learn this knowledge and relationships purely from large-scale data. 
In other words, token representations require FMs to rediscover this knowledge that humans have already formalized mathematically. 
For example, AlphaGeometry had to rediscover millions of theorems and proofs that were already known to geometry literature \cite{trinh2024solving}.
Such the learning-from-scratch paradigm not only demands massive data and computation but also can fail to capture the underlying physical relationships that might not be able to infer directly from finite given training samples \cite{firoozi2023foundation,gupta2024essential}.
We assume that FMs building upon DT representations can achieve higher-level reasoning by explicitly incorporating world knowledge into the representation rather than forcing them to learn from scratch as statistical patterns.

\paragraph{Limitations of Token Representation}
Current enhancement of real-world knowledge understanding in vision-language FMs like InternLM-XComposer2-4KHD \cite{dong2024internlm} and Phi-3 Vision \cite{abdin2024phi} mainly depends on adding the number of vision tokens obtained from multiple views. 
However, theoretical and empirical analyses reveal its limitation in terms of the scaling \textit{i}.\textit{e}., the performance $S(N)$ with respect to the number of vision tokens $N$ follows a weak scaling behavior $S(N) \approx (c/N)^\alpha$, where $c$ and $\alpha$ are parameters \cite{li2024scaling}. 
It yields that increasing the quantity of vision tokens cannot overcome the limitations in capturing high-level world knowledge such as semantic relationships, due to the mismatch between discrete tokenization and the continuous nature of real-world interactions.
%
%
For example, in a video of a ball rolling down an inclined plane, tokenization can reduce the continuous motion into a sequence of discrete snapshots, which thus not only discards information about velocity, acceleration, and gravitational forces, but also forces the FM to learn reconstructing these well-established physical relationships from scratch \cite{firoozi2023foundation}.
Consequently, FMs operating on token representations can struggle to achieve generalization. 
Rather than directly leveraging existing mathematical-formulated understanding of physics and geometry as can be achieved in DT representation, token representations rely on pattern matching that may only capture surface-level correlations. 
%

\paragraph{Encode Real-World Knowledge and Relationships into DT Representation}
DT representations can address these limitations by explicitly encoding real-world knowledge and domain-specific constraints through mathematically precise relationships tailored to specific domains and tasks. 
It can be achieved through a layered representation framework where geometric relationships are encoded as spatial transforms and semantic knowledge as graphs with defined predicates and relations \cite{ding2024digital}.
It therefore can reduce the amount of data needed for training, as FMs no longer need to learn all knowledge from scratch \cite{kannapinn2024twinlab}. 
Another advantage of DT representations lies in their ability to integrate established domain knowledge \cite{kapteyn2021probabilistic}. 
For example, in robotics applications, Ding \textit{et al.}~\cite{ding2024towardsR} extracted digital twin representation to represent the status of peg transfer task and provide this representation to a LLM-based agent with prior knowledge of peg transfer task to enable flexible and intractable long-horizon planning.
%
%
Such structured integration of domain expertise enables FMs to maintain physical consistency even in complex scenarios, which therefore stands in stark contrast to token representation where they might fail to capture subtle interactions.

\paragraph{Practical Benefits and Impact}
The advantages of knowledge-aware DT representations manifest across diverse real-world applications of FMs. 
Consider autonomous robotics tasks that require complex interaction with the physical environment \cite{stkaczek2021digital}. 
%
%
In contrast, a DT representation can explicitly encode fluid behavior, container geometry, and physical constraints, enabling direct reasoning about pouring actions while maintaining conservation of mass \cite{balazadeh2024synthetic,khamkar2024digital}. 

\subsection{Assumption 2: DT Representation Enables More Practical Data Synthesis at Scale}

Scaling FMs in different ways such as training and inference is receiving more attention in both industry and academia \cite{li2025minimax}. 
Previous works demonstrate that increasing model size and training data scale contribute to improvement in model capabilities \cite{kaplan2020scaling}.
This scaling pattern relies on access to massive and high-quality training data.
However, acquiring real-world multimodal data at scale suffers from high costs, time-intensive collection processes, privacy concerns, and difficulty in capturing rare but critical scenarios. 
In medical imaging, for instance, training multimodal diagnosis FMs can require management of sensitive patient information while ensuring consistency across diverse data types like CT scans and clinical records \cite{gehrmann2023prevents}.
Data synthesis has become a widely adopted solution in response to the scaling challenges.

\paragraph{Limitations of Token Representation}
Existing data synthesis methods primarily rely on generative models or rule-based simulations. 
%
%
However, generative models trained on token representations fail to capture the ``long tail'' distribution of real-world scenarios beyond superficial patterns \cite{manduchi2024challenges}. 
This can lead to generating data that violates physical principles and domain constraints \cite{shumailov2024ai}. 
For example, in molecular biology simulations, token-based generative approaches struggle to maintain valid chemical structures \cite{zeng2022deep}.
%
%
Traditional rule-based simulations attempt to address this but often oversimplify complex interactions. 
For instance, in biomolecular systems, rigid simulation frameworks fail to capture the nuanced site-specific dynamics of protein interactions \cite{chylek2015modeling}.

\paragraph{Data Synthesis with DT Representation}
DT representations can transform the paradigm of data synthesis by leveraging the semantic richness and physical awareness inherent in the corresponding DT with respect to the DT representation \cite{jagatheesaperumal2023semantic}.
In specific, DT provides complete virtual replications of physical entities that mirror their real-world properties and behaviors while DT representations serve as the underlying representation layer that encodes these properties in a format suitable for FMs. 
This separation enables to synthesis of more practical data at scale.
By constructing DT representations that capture essential characteristics like geometry, physical properties, and domain-specific constraints, we can generate synthetic data that maintains high fidelity to real-world behavior. 
Unlike approaches based on token representation that rely purely on learned statistical correlations, DT representations explicitly encode physical constraints and domain knowledge through their outcome-driven design. 
This encoding preserves both intra-modality relationships (\textit{e}.\textit{g}., spatial and temporal dynamics) and inter-modality interactions (\textit{e}.\textit{g}., how visual changes correspond to physical state changes) \cite{sigawi2023using}.
Moreover, the physics-aware nature of DT representations enables the exploration of edge cases while maintaining physical validity. 
For example, in the domain of autonomous driving, the TWICE dataset uses DT representations to generate physically accurate simulations of adverse weather conditions \cite{neto2023twice}.

\paragraph{Bridging the Sim-to-Real Gap}
DT representations can also bridge the performance deviation between simulated applications and their real-world deployment \cite{mcmanus2024effects}. 
By maintaining consistent physical and semantic properties across simulated and real environments, FMs trained on DT-based simulations transfer more effectively to real-world applications. 
The ACDC demonstrates this through its use of ``digital cousins'' \textit{i}.\textit{e}. DT representations that share geometric and semantic affordances with real-world objects, to enable zero-shot deployment of trained policies in real-world scenarios \cite{dai2024acdc}.
%

\subsection{Assumption 3: DT Representation Contributes to Better Causal Reasoning}

Causal reasoning, \textit{i}.\textit{e}. the ability to understand and infer cause-and-effect relationships enables humans to not only recognize correlations between events but also understand the underlying mechanisms that drive these relationships, predict future outcomes, and reason about hypothetical scenarios. 
In the context of FMs, causal reasoning encompasses three capabilities, namely identifying cause-and-effect relationships (interventional reasoning), understanding why specific outcomes occur (attributional reasoning), and predicting outcomes under hypothetical conditions (counterfactual reasoning) \cite{yang2024critical}. 
Current FMs struggle with these aspects of causal understanding, as LLMs achieve almost random performance on pure causal inference tasks \cite{jin2023can}.
%
%
This can potentially result in hallucination, where FMs generate plausible but factually incorrect information, and spurious correlations, where FMs learn superficial patterns that don't reflect true causal relationships \cite{li2024look}. 
We argue that DT representations offer one direction in addressing these limitations by explicitly encoding causal mechanisms and physical constraints.

\paragraph{Limitations of Token Representation}
Another challenge of token representation in FMs lies in their reliance on statistical learning to infer causal relationships. 
These FMs identify patterns and correlations but cannot distinguish genuine causal relationships from mere statistical associations, which stem from two factors \cite{wu2024causality}. 
Firstly, token representations force FMs to learn causal relationships solely from observational data, which violates the fundamental principle in causal inference that correlation does not imply causation \cite{willig2022can}. 
Second, the discretization process of tokenization fragments continuous causal chains into disconnected sequences, effectively destroying important information about the mechanisms that link causes to effects \cite{schmidt2024tokenization}. 
These limitations become more problematic in counterfactual reasoning tasks, where FMs must evaluate hypothetical scenarios that deviate from observed patterns \cite{wang2024evaluating}. 
FMs with token representation struggle to generate physically plausible counterfactuals because they lack explicit representations of the causal mechanisms that govern real-world phenomena \cite{wu2024semantic}. 
It not only affects the FM's ability to perform reliable causal reasoning but also contributes to broader issues like hallucination and poor generalization to novel scenarios \cite{wang2024evaluating}.

\paragraph{Enhanced Causal Understanding through DT Representations}
%
Unlike token representation where FMs need to learn causal relationships from scratch, DT representations can explicitly encode causal mechanisms and relationships into their structure \cite{jakovljevic2022towards}. 
This can be achieved through three mechanisms.
First, DT representations maintain explicit cause-and-effect relationships through their outcome-driven design. 
By preserving the continuous nature of physical processes and their underlying mechanisms, DT representations enable FMs to reason about causality in a more principled way \cite{kapteyn2022data}. 
Second, DT representations enable counterfactual reasoning by maintaining physically valid state spaces. 
When generating counterfactuals, FMs can leverage the explicit physical constraints and domain knowledge encoded in DT representations to ensure their predictions remain consistent with real-world causal mechanisms \cite{dai2024acdc}. 
Third, DT representations facilitate interpretable causal attribution by maintaining clear relationships between components and their effects. 
Unlike the opaque relationships learned in token representations, DT representations provide transparent causal chains that can be traced and verified in causal diagrams \cite{balu2022physics,somers2022reliable}. 
This transparency helps reduce hallucination by grounding FMs' predictions in explicit causal mechanisms rather than learned statistical patterns.

\paragraph{Practical Impact on Model Reliability}
The enhanced causal reasoning capabilities enabled by DT representations can directly address several challenges in current FMs. 
More notably, they can potentially help reduce hallucination by grounding FM predictions in explicit causal mechanisms rather than learned statistical patterns. 
When the FM with DT representations makes predictions, it must do so under the encoded physical constraints and causal relationships, reducing the likelihood of generating physically impossible or causally inconsistent outputs.
For example, in autonomous driving scenarios, DT representations enable more reliable reasoning about the potential consequences of different actions by maintaining explicit causal models of vehicle dynamics and traffic interactions \cite{neto2023twice}. 
Similarly, in medical diagnosis applications, DT representations can help models reason about disease progression and treatment effects by encoding known causal relationships between symptoms, conditions, and interventions \cite{gehrmann2023prevents}.
This improved causal reasoning capability also enhances model robustness to distribution shifts. 
For example, previous work show causal structure helps create semantic representation with DT that is causally invariant, helping generalize the learned knowledge to unseen scenarios \cite{thomas2023causal}.
Because DT representations encode causal mechanisms rather than surface-level correlations, FMs can better generalize to novel scenarios by reasoning from first principles rather than relying on pattern matching.

\subsection{Assumption 4: DT Representation as a Unified Semantic Representation}

A unified semantic representation aims to provide a common semantic space where different modalities, whether visual, textual, temporal, or physical, can be encoded and processed while maintaining their natural relationships and constraints. 
However, current multimodal FMs lack such unified representation.
They rely on complex architectures that artificially bridge different modalities through learned statistical correlations, which thus fail to capture the inherent connections between modalities, leading to poor generalization. 
We argue that DT representations can offer a natural and elegant solution by providing a unified semantic representation grounded in physical reality.

\paragraph{Limitations of Token Representation}
Token representations fail to achieve a unified semantic representation necessary for multimodal understanding \cite{qu2024tokenflow}. 
Token representations can create artificial boundaries between modalities by discretizing continuous information into separate visual, textual, and temporal tokens. 
This forced separation requires FMs to learn cross-modal relationships purely through statistical correlation, rather than preserving the natural semantic connections that exist in the real world. 
Moreover, these representations lack a shared semantic foundation across modalities, for example, visual tokens capture low-level pixel patterns while text tokens encode linguistic structures, with no inherent mechanism to maintain their interconnections. 
For example, previous work pointed out that visual patches and text tokens differ in semantic levels and granularities, making direct alignment challenging \cite{chen2023revisiting}.
%
%

\paragraph{Unified Semantic Space Through DT Representation}
DT representations provide a unified semantic representation that naturally preserves relationships across modalities and domains. 
Rather than forcing different modalities into token sequences, DT representations maintain a continuous semantic space grounded in real-world properties and constraints. 
This can be realized upon previous work in neural scene representations, where methods like Neural Radiance Fields (NeRFs) demonstrate their capabilities of unified 3D representations in maintaining consistency across multiple views and modalities \cite{mildenhall2023nerf}. 
It can reduce domain gaps by encoding information through shared physical and geometric properties rather than learned statistical correlations between discrete tokens \cite{segovia2022design}.

\paragraph{Enhanced Generalization Through Unified Representation}
The unified semantic nature of DT representations directly contributes to improved generalization capabilities in FMs. 
%
%
To be more specific, DT representations can encode physical and semantic principles that remain consistent across domains. 
This inherent structure enables FMs to generalize more effectively to novel scenarios by reasoning from first principles rather than relying on learned patterns \cite{sudhakar2023exploring}.
For example, Ding \textit{et al.}~\cite{ding2024towardsS} map video clips of cholecystectomy surgery into unified digital twin representation with tissue and surgical tool masks with corresponding geometry, which can enhance the FM's generalization when applied to out-of-distribution test samples.
Another paper \cite{shen2025online} introduces an agent-based framework that decouples perception and reasoning through a just-in-time digital twin representation to perform video reasoning segmentation, which also demonstrates how DT representations effectively preserve semantic, spatial, and temporal relationships in video compared to its token representation counterparts.
Its follow-up \cite{shen2025operating} applies these DT representations to the analysis of operating room workflow efficiency, creating DT representations that preserve both semantic and spatial relationships between the OR components for FM to reason.

\subsection{Assumption 5: DT Representation Can Improve the Transparency and interpretability}

Transparency refers to the ability to understand how an AI model processes data and arrives at its decisions.
Interpretability, on the other hand, focuses on making these processes and decisions understandable to humans in terms of familiar concepts and relationships. 
The increasing complexity and impact of FMs make transparency and interpretability more important than ever by enabling users to verify, trust, and deploy FMs in real-world applications. 
To be more specific, they enable developers and users to verify that FMs are functioning as intended and following necessary safety constraints. 
They also allow the identification and correction of biases or errors in reasoning. 
However, token representations limit transparency and interpretability in different ways.
We argue that DT representations offer a natural framework for building more transparent and interpretable FMs.

\paragraph{Limitations of Token Representation}
FMs using token representations operate largely as black boxes, making it challenging to understand and verify their decision-making processes. 
The major issue lies in how token representations abstract information into high-dimensional spaces that lack clear semantic grounding and are thus difficult to interpret \cite{robinson2024structure}. 
Previous work also pointed out that not every point in the ambient latent space corresponds to a token, highlighting the abstract nature of these representations on the other side \cite{viswanathan2025geometry}.
When processing physical interactions and multimodal data, these representations fragment continuous phenomena into discrete tokens, obscuring the natural relationships and constraints that govern real-world behavior. 
For example, previous work shows that continuous features generally outperform discrete tokens, particularly in tasks requiring fine-grained semantic understanding due to the absence of token granularity and inefficient information retention \cite{wang2024comparative}.

\paragraph{Enhanced Interpretability Through DT Representation}
DT representations address these transparency challenges by maintaining explicit physical and semantic meaning throughout the processing pipeline \cite{katsoulakis2024digital}.
Unlike token representations that embed information in abstract spaces, DT representations preserve coherent relationships between physical properties, geometric constraints, structural properties, and causal interactions \cite{lammini2022geometric}. 
This preservation of meaningful structure enables direct interpretation of the model's internal states and reasoning processes at multiple levels of abstraction \cite{siyaev2023interaction}.
At the representation level, DT representations maintain physical quantities in their natural units and relationships. 
At the reasoning level, DT representations enable step-by-step examination of the FM's decision process through physically grounded intermediate states. 
When planning complex actions, users can trace how the FM progresses from initial observations to final decisions through a series of meaningful physical and semantic transformations.


\section{Conclusion}
This position paper has argued that the machine learning community should seriously consider DT representations as an alternative to current FMs with token representations. 
We present discussions on how DT representations can address the limitations of token representations across multiple dimensions, namely capturing real-world knowledge and relationships, enabling practical data synthesis, supporting causal reasoning, providing unified semantic representations, and improving transparency.
Meanwhile, several future research directions emerge such as the development of efficient frameworks for constructing and maintaining DT representations at scale, theoretical foundations for understanding their properties, specialized evaluation benchmarks, and hybrid architectures that can combine the strengths of both DT and token representations.

\bibliographystyle{plain}
\bibliography{main.bib}

\end{document}